# Bio-inspired Multi-robot Autonomy

S. Chirayil Nandakumar, S. Harper, D. Mitchell, J. Blanche, T. Lim, I. Yamamoto and D. Flynn
*Member, IEEE[1]*

*Abstract*—Increasingly, high value industrial markets are driving trends for improved functionality and resilience from resident autonomous systems. This led to an increase in multi-robot fleets that aim to leverage the complimentary attributes of the diverse platforms. In this paper we introduce a novel bio-inspired Symbiotic System of Systems Approach (SSOSA) for designing the operational governance of a multi-robot fleet consisting of ground-based quadruped and wheeled platforms. SSOSA couples the MR-fleet to the resident infrastructure monitoring systems into one collaborative digital commons. The hyper visibility of the integrated distributed systems, achieved through a latency bidirectional communication network, supports collaboration, coordination and corroboration (3C) across the integrated systems. In our experiment we demonstrate how , an operator can activate a pre-determined autonomous mission and utilize SSOSA to overcome intrinsic and external risks to the autonomous missions. We demonstrate how resilience can be enhanced by local collaboration between SPOT and Husky wherein we detect a replacement battery, and utilize the manipulator arm of SPOT to support a Clearpath Husky A200 wheeled robotic platform. This allows for increased resilience of an autonomous mission as robots can collaborate to ensure the battery state of the Husky robot. Overall, these initial results demonstrate the value of a SSOSA approach in addressing a key operational barrier to scalable autonomy – resilience*. Keywords- Multi-robot system, Cooperating robots, Human Factors and Human-in-the-Loop.*

## I. INTRODUCTION

Robots are moving beyond short-term, limited task missions within well-defined environments specifically optimized for robotic and autonomous operations [1]. As the global energy markets transition to net-zero energy infrastructure, continued growth will rely on significant advancements in scalability and capability of multi-robot fleets operating in real-world conditions. These conditions can be characterized as highly variable, widely distributed, high-risk environments, such as offshore wind installations and nuclear power plants [2]. To create productive, resilient and safety compliant MR-fleets, we must address central barriers to the generalization of AI. Today, many barriers exist due to the deterministic design methodologies that aim to optimize well-defined objectives, assuming real missions either replicate offline training environments or conservative training scenarios. Also, it assumes that all objectives are well understood and appropriately prioritized throughout the lifecycle of the mission. Due to this, AI has the inability to adequately rationalize large amounts of additional and highly diverse dynamic information. It also constrains autonomy to the pursuit of predefined objectives, potentially irrelevant or erroneous, when operating in uncertain and changing circumstances. Barriers are also created by the inability to combine human phronetic knowledge, distributed data, knowledge bases, symbolic and statistical reasoning. In this paper we focus on operational barriers, namely run-time safety compliance, reliability, resilience, and optimization of human and multi-robot fleet (MR-fleet) collaboration, for scalable autonomy. Consequently, we identified the use of multi-robot systems in conjunction with Digital Twins (DT) as shown in Fig 1, as a pathway to advance global energy digitalization [3, 4].

This paper demonstrates three missions with increasing interactions across robots as utilizing Cooperation, Collaboration and Corroboration ($C^3$ governance) within a SSOSA.

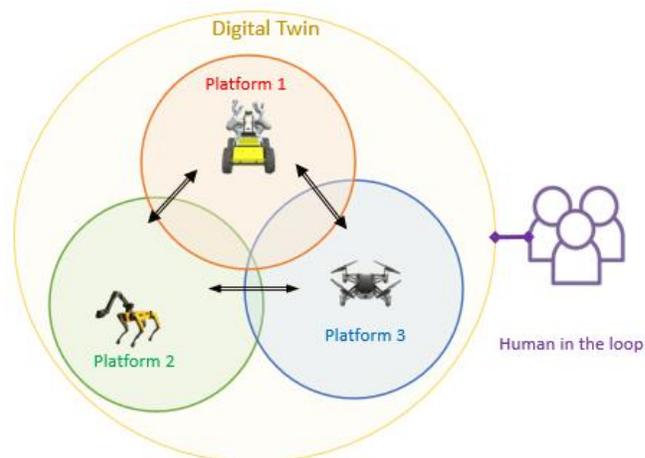

Figure 1. Bio-inspired SSOSA for multi-robot autonomy where double headed arrows represent bidirectional communication.

[1] *Research supported by the Offshore Robotics for the Certification of Assets (ORCA) Hub under EPSRC Project EP/R026173/1 and MicroSense Technologies Ltd in the provision of their patented microwave FMCW sensing mechanism (PCT/GB2017/053275).

S.Nandakumar, S. Harper ,D. Mitchell, J. Blanche and T. Lim are with the Smart Systems Group at Heriot-Watt University, Edinburgh, EH14 4AS (email: sc2039@hw.ac.uk; Sam.Harper@hw.ac.uk; dm68@hw.ac.uk; J.Blanche@hw.ac.uk; T.Lim@hw.ac.uk.

I.Yamamoto was with Nagasaki University, Nagasaki, 852-8521, Japan (email: iyamamoto@nagasaki-u.ac.jp)

D.Flynn is with James Watt School of Engineering, University of Glasgow, UK, G12 8QQ (email: David.Flynn@Glasgow.ac.uk).

AI models are trained to detect a battery box and Husky robotic platform through any of the five cameras of the Spot Robot enabling symbiotic interactions across robots and environment. The mission scenarios validate improved resilience in a multi-robot autonomous missions where the mission evaluation can be viewed via [5].

This publication is structured as follows. In Section II, a comparative analysis of multi-robot system strategies is carried out, which identifies ours as the most scalable and flexible. Section III presents the bio-inspired novel methodology, SSOSA and $C^3$ governance for Symbiosis, the digital twin enabling symbiosis, and improved resilience via symbiosis, enhancing multi robot autonomy . An overview of the experimental setup and summary of three missions are provided in Section IV. The results are discussed in Section V, where symbiosis via $C^3$ governance enhancing resilience in multi robot autonomous missions is validated, followed by the conclusion in Section VI.

## II. RELATED WORK

This section explores the definition of symbiosis in nature, and provides a review of current multi-robot system strategies alongside comparative analysis of current multi robot systems. Symbiosis in nature is well defined in [6, 7], and includes relationships that can provide benefits for each of the participating organisms, such as pollination. This represents a mutually beneficial symbiotic relationship.

The developments in offshore robotics currently follow a product-centric pathway, however, there is an increasing trend of coupling different robotic systems to create more advanced solutions for inspection and maintenance tasks [3]. A trend which exists in the offshore sector includes residential multi-robot systems to address new operations and maintenance challenges [8, 9].

In conducting this literature review, a keyword search was completed on google scholar for the following searches: '*Multi-robot systems*' and '*Bio-inspired multi robot systems*'. The resulting research findings were evaluated for $C^3$ and heterogeneous properties, where researchers use several strategies related to artificial intelligence and bio-inspired interactions to develop multi-robot autonomous missions.

Yi *et al.* demonstrated a bio-inspired task assignment solution using a Self-Organizing Map (SOM) based approach for multi-robot systems, which can provide path planning and overcome situations with high uncertainty, such as robots added or removed from the fleet [10]. A key observation includes the assumption that the robots within the robotic swarm are of same capabilities (same robot) and omits the inclusion of interactions among heterogenous robotic systems.

Similarly, Lima *et al* [11] developed a Hybrid Cellular Automata Ant Queue (HCAAQ) model capable of adapting the current dynamics of the multi-robot mission if a robotic agent faces significant detriment to the robot platform. The HCAAQ model allows for awareness in the number of available robotic platform in the multi robot mission. There are advantages present within HCAAQ verses SOM due to the ability for collaboration between robots and the ability to create more advanced mission plans. A disadvantage of both approaches includes that the systems were only tested under simulation.

A multi-robot task scheduling named Ant-Colony Optimization (ACO) algorithm utilized the idea of pheromone in ants by observing the following: Ants leave pheromones as they leave in a path , Ants will be more likely to follow a path with more pheromones nearby. This idea was used to create task allocation strategies in robotic missions. However, the test was demonstrated only in simulation and they assumed all the robots are having exact same capabilities [12].

Ranjbar-Sahraei *et al.* demonstrated algorithm Stigmergic Coverage (StiCo) inspired from ants and the idea of pheromones for a multi-robot mission. This task allocation algorithm based on stigmergy perform well on robot swarms, where robots have same features, and with heterogenous robotic systems, where the robots have varied capabilities [13]. Moon *et al.* demonstrated the use of Programmable Logic Controller (PLC) based coordination schemes for multi-robot systems with varied capabilities [14]. However, they lack the essential resilience as there are no strategies to identify and overcome uncertain situations required for robots to perform a fully autonomous multi-robot mission. An overview of the active research in the multi-robot field is tabulated in table I.

This literature review has utilized $C^3$ governance and the integration of heterogeneous robots as key limitations in bio inspired multi robot autonomy. Understanding the need for autonomy of multi-robot systems for the long term sustainability and reliability of missions, this review highlights an opportunity via a SSOSA [6] via $C^3$ governance utilizing the unique capabilities of heterogenous robots to create symbiosis, thereby enhancing the resilience of the multi-robot autonomous missions.

TABLE I. MULTI ROBOT SYSTEM STRATEGIES

| *Approach* | *Cooperation* | *Collaboration* | *Corroboration* | *Heterogenous Robot Integration* |
|---|---|---|---|---|
| SOM [10] | ✓ | ✗ | ✗ | ✗ |
| HCAAQ model [11] | ✓ | ✓ | ✗ | ✗ |
| ACO model [12] | ✓ | ✗ | ✗ | ✗ |
| StiCo [13] | ✓ | ✓ | ✗ | ✗ |
| PLC based Coordination [14] | ✓ | ✗ | ✗ | ✓ |
| SSOSA [6] | ✓ | ✓ | ✓ | ✓ |

## III. METHODOLOGY

### A. Symbiotic System of Systems Approach and $C^3$ governance

The offshore industry use robotic systems to read gauges, detect leaks, scan the horizon, take video and audio, and identify anomalies [9]. This results in robots with different capabilities being used in offshore sector and the optimization of fleet performance will require interconnection of robotic agents. Interestingly, the Symbiotic relationships in the natural environment can guide us in developing self-sustaining multi-robot systems, as proposed by Mitchell *et al* [6]. Specifically, we proposed a Symbiotic System of Systems Approach (SSOSA) wherein the Symbiotic interactions define the informal and formal relationships that operate under $C^3$ governance. $C^3$ governance represents the systems capable of "Collaboration, Cooperation and Corroboration" [15]. Similar to nature's symbiosis, SSOSA demonstrate and defines various types of Symbiosis possible in the multi- robot systems[6].

### B. Symbiosis of multi-robot fleets via digital environment

The concept of Symbiosis in a multi-robot system is enabled using a digital environment, also known as DT. Here, DT defines the digital representation of the Symbiotic System of System Approach with bidirectional communication capability enabling multi-robot interactions as displayed in Fig 2 . The DT can capture data from the robotic systems for remote visualization, analysis and to corroborate with the Human-In-The-Loop (HITL). Moreover, it transfers data to the robotic system, commanding robots to execute symbiotic missions or schedule an autonomous inspection or maintenance task. Thus, DT can act as the intermediary between robots or an environment that collects multiple robot data and triggers symbiotic features such as Mutualism, Commensalism and Parasitism using the $C^3$ governance strategies of Cooperation, Collaboration and Corroboration during the multi-robot autonomous missions.

### C. Enabling Multi-robot Mission resilience through Symbiosis

Mission resilience is the capability of the system to understand and overcome any uncertain circumstances during the mission. And the major element to long term multi-robot autonomy is the mission resilience. Thus, improving mission resilience through path and mission planning algorithms for multi-robot systems is an active area of research. This work explores and demonstrates how the SSOSA can enhance multi-robot mission resilience. Specifically, we demonstrate how the corroboration of robotic systems via another robot enhances the overall system's safety and trustworthiness. And then, use the Collaboration and Corroboration aspects of $C^3$ governance to improve the mission resilience by demonstrating how an autonomous robot platform can help in mitigating an issue encountered by a nearby heterogenous robot. Thus, paving the pathway to the long-term self-sustenance of multi-robot autonomous systems.

## IV. DEMONSTRATION

### A. Overview of the experimental setup

An autonomous mission was created to verify the effectiveness of a bio-inspired multi-robot fleet. To validate the methodology, several off-the-shelf robots were utilized where wireless communication were established to allow for an operational overview via the DT. The demonstration also included several security cameras which were directly connected to the DT via USB and were used for corroboration of the multi-robot mission. The types of robotic platforms used are displayed in Table II.

The offshore multi-robot mission scenario is reconstructed in the laboratory settings, as shown in the fig 3. The missions for the demonstration are

- M1: Systems check
- M2: Autonomous Beyond the Visual Line of Sight (BVLOS) mission
- M3: Autonomous Battery replacement mission

These missions demonstrate increasing complexity and interconnectedness between robotic systems and how the various features of $C^3$ governance are creating mission resilience. The details of the different mission execution are provided in the subsequent sub-sections.

TABLE II. ROBOTIC PLATFORMS USED IN DEMONSTRATION

| Robot Platform | Name | Specification | Use cases |
|---|---|---|---|
| 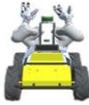 Platform 1 | Clearpath Husky with dual UR-5 and Robotiq grippers. | Wheeled robotic platform with obstacle avoidance capability equipped with LIDAR, two manipulators with 3-fingered grippers and a depth camera. | Transportation of goods, inspection and maintenance of the offshore environment. |
| 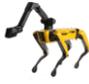 Platform 2 | Boston Dynamics Spot with Arm. | Quadruped robotic platform with obstacle avoidance capability. A manipulator is attached with a two-finger gripper. The platform also has 5 Depth cameras on the base and a color camera in the Arm. Application development is done using Boston Dynamics Python SDK | Walking, climbing staircase, opening doors, inspection and maintenance of the offshore environment. |
| 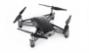 Platform 3 | DJI Ryze Tello EDU | A quadcopter with IR sensors for height measurement and camera for video capture. | Inspection in the air, manoeuvring through narrow air spaces. |

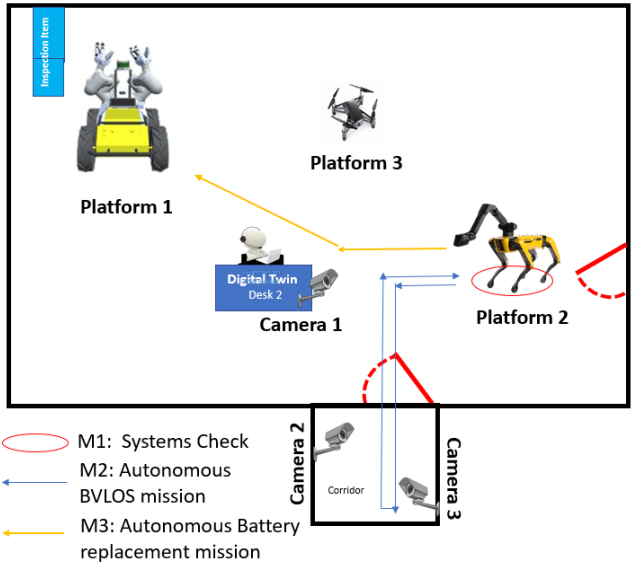

Figure 2. The three offshore mission scenarios in the laboratory environment.

The digital environment is developed in Unity as shown in Fig 3. The bidirectional communication between the robotic platforms, security cameras and Unity environment is established by creating server and client scripts. Thus, enabling the telemetry data streaming from the Spot, Husky and Tello drones. Moreover, HITL can trigger autonomous mission scripts on Boston Dynamics Spot Robot via DT due to the bi-directional feature. The script for communication in Unity is developed C# language. The scripts for Spot are written in python utilizing the Python SDK and the documentation provided by the Boston dynamics[16].

B. *Summary of Autonomous Missions*

We applied the bio-inspired methodology to the following missions to validate that $C^3$ governance enhances symbiosis and lead to increased safety, reliability and resilience in the multi-robot missions.

*1) Mission 1: Systems check of platform 2 with corroboration from platform 3*

In mission M1, the corroboration aspect of the $C^3$ framework is demonstrated by the HITL and Platform 3 corroborating the autonomous full body movement of the Platform 2 visually via reading the real time telemetry data from Platform 2 and Video stream from Platform 3, to check and confirm if there are any issues with the hardware of the platform. The sequence of events during the mission occurred as within Fig 4. Specifically, as the autonomous mission is triggered, the motors in the platform 2 will be turned on and the quadruped platform will be in standing position. Then, the arm of the platform is unstowed. Further, the robotic arm will try to draw a circle of radius 1 metre infront of the robot while the quadruped robot itself will be doing a walking along an imaginary square on the ground of side length 1 metre. This shows that the robot is capable of moving its arm and legs simultaneously, and all the movements are corroborated via the platform 3 which is teleoperated by HITL. The interactions across the robots, DT and HITL are displayed for each sequence of events in Fig 5.

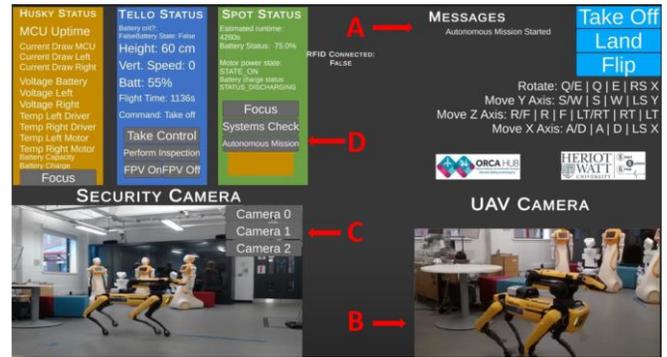

Figure 3. The Digital Twin interface developed in Unity **A**: Spot notifies regarding the mission status. **B**: Corroboration via DJI Tello Drone. **C**: Corroboration via security cameras. **D**: Tabs to trigger autonomous missions on Spot.

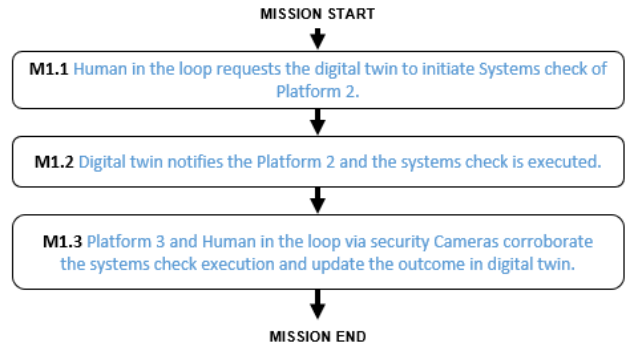

Figure 4. Sequence of events for mission.

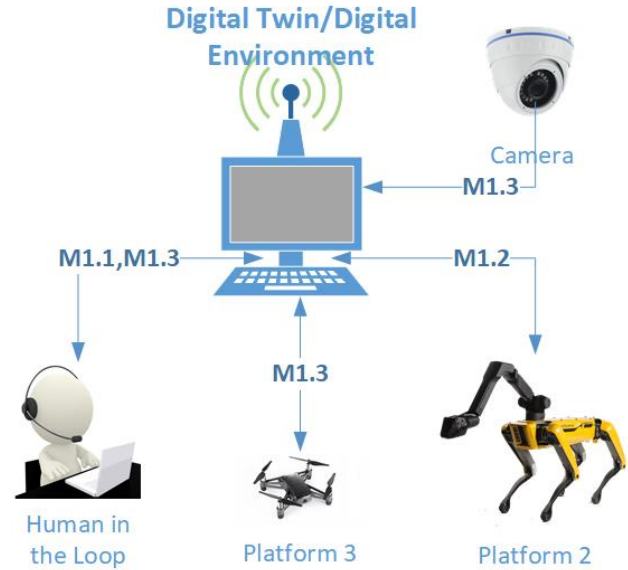

Figure 5. Mission M1: Systems check.

*2) Mission 2: Autonomous BVLOS mission.*

In the M2, the effectiveness of the corroboration with multiple robots and other sensor feedbacks such as a Security Camera is emphasized via the execution of a fully autonomous BVLOS mission. Fig. 4 shows the view from DT during the mission M2 where the HITL press a tab located near the 'D' in the Fig. 3. The sequence of events during the mission occurred as within Fig 6.

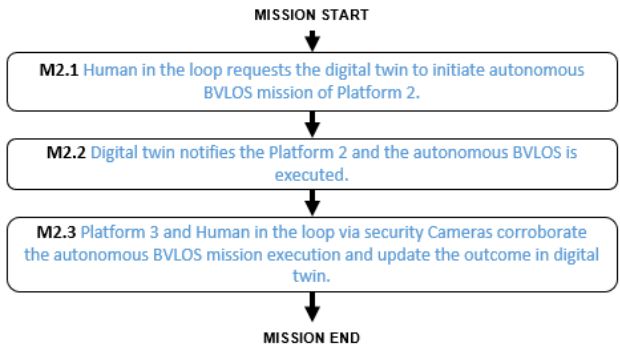

Figure 6. Sequence of events for Mission M2.

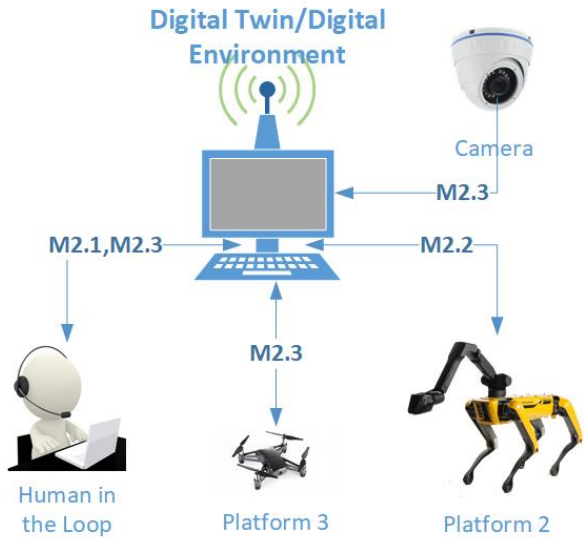

Figure 7. Mission M2: Autonomous BVLOS Mission.

The interactions across the robots and DT is displayed for each sequence of events in Fig. 7. The fully autonomous mission was developed using the "Autowalk" feature in the python SDK of the Platform 2. The

Autonomous mission python script is developed and placed in a PC connected to the Platform 2. This script is triggered from the Digital Twin in Unity running on MacBook Pro by developing a Server node in Unity and a Client node in a PC running on Ubuntu 18.04 connected to platform 2. All the communication is using the Wifi. Moreover, Platform 3 is teleoperated by the HITL. The details of the mission execution are shown in the figures below.

*3) Mission 3: Autonomous Battery replacement scenario during a Multi-Robot Mission*

This demonstration showcases how a robotic platform can help another robotic platform during a mission-critical condition. In the mission scenario, Platform 1 is out of battery and could not continue the mission until a replacement battery is received. The Platform 1 notifies the Digital Twin and the Digital Twin via the HITL triggers the Autonomous Battery search Mission M3. In the Mission, the Platform 2 search for a battery box in the environment, collects the battery box, search for the Platform 1 in the surrounding and carry the battery to place it near the platform 1. This battery box can be used by the platform 1 and enable mission continuity, thus improving multi-robot mission resilience.

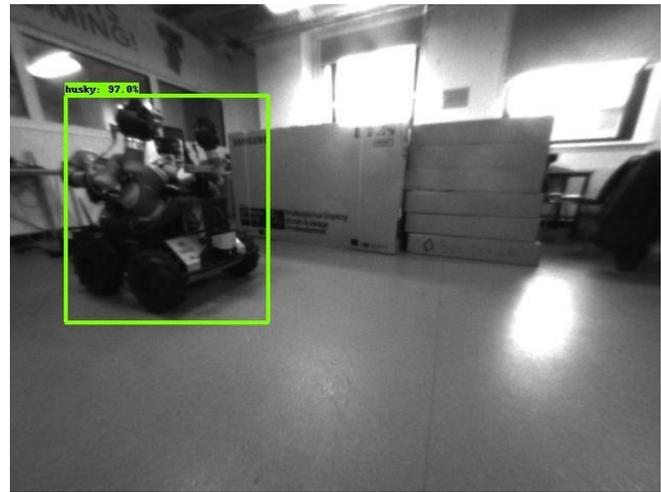

Figure 8. AI model trained to detect the Husky robot. Specifically, for this image, the trained model is 97% confident that the object detected is Husky. The image shown is the view from the front left camera of the Boston Dynamics Spot Robot.

Specifically, Platform 2 autonomously search via its five cameras, the battery box in the environment. Once the battery is detected with higher than 60% confidence by the AI model, the Platform 2 autonomously plan the motion utilizing Boston dynamics APIs to reach the battery box and grasp it. Once the grasping is successful, the Platform 2 continuously search via its five cameras for the Platform 1 in the surrounding. Once Platform 1 is detected with 60% or higher confidence by the AI model as seen in fig 8, platform 2 autonomously plan the motion, then carries and places the battery box to two meters in front of the Platform 1. Fig 9 shows an image captured during the mission M3.

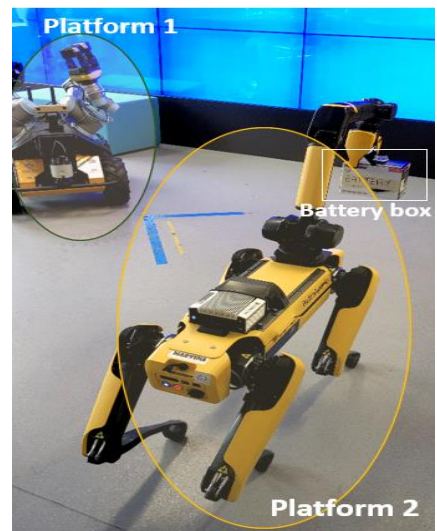

Figure 9. An image captured during the M3 mission. Here, Platform 2 has already identified the box, grasped it, identified the Platform 1, and moved towards the Platform 1 to give the battery box. All the tasks are fully autonomous.

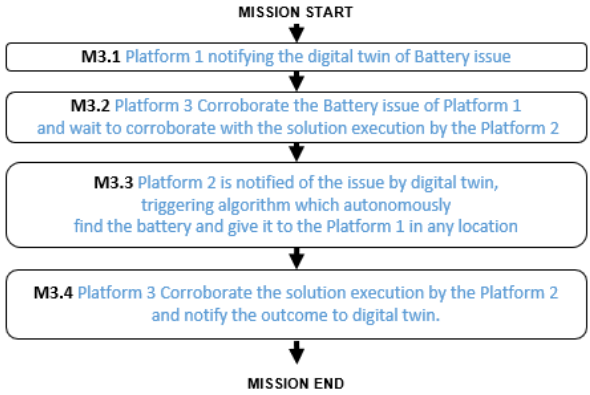

Figure 10. Sequence of Events demonstrating Symbiosis.

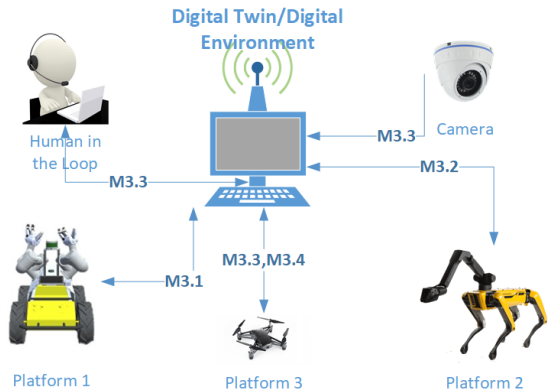

Figure 11. Autonomous battery replacement mission

The sequence of events during the mission occurred as within Fig 10. The interactions across the robots and DT are displayed for each sequence of events is in Fig 11.

For the M3 mission, Two AI models for object detection were used. AI models defined in TensorFlow were trained on an Ubuntu 18.04 laptop with an NVIDIA GTX 1060 3GB Graphics card and 16 GB RAM. The 560 and 845 images containing the object of interest, the "battery pack" and "Husky robot", were collected in equal numbers from the five cameras of the Spot robot. The five cameras are seen in front-left, front-right, left-side, right-side and back of the Spot. The images were preprocessed and labelled using labelling software. Transfer learning was used to train a pretrained model SSD ResNet50 V1 FPN 640x640 to learn to detect these specific objects. Both Models learned to detect items satisfactorily as the training loss was less than 0.03 after 20,000 Epoch of training.

The Autonomous walking and grasping using the arm are done using the Boston Dynamics API, as demonstrated in the tutorials of the Boston Dynamics SPOT SDK documentation[16]. The python scripts developed for autonomous search for battery, grasping and autonomous search for Platform1 and walking towards it, and the AI models for battery and Platform 1 detection are triggered by the Digital Twin when there is a battery issue in a nearby Platform 1.

## V. DISCUSSION

The above three missions validate that symbiosis via strategies of $C^3$ governance can improve autonomous multi-robot mission resilience. Specifically, for M1, corroboration via Platform 3 enhanced the trust in the system check mission execution. In M2, during the BVLOS mission, having Platform 3 cooperate and corroborate throughout and HITL corroborating with security cameras increased reliability of the mission progress. Furthermore, finally, in M3, the fully autonomous collaboration between platform 2 when there was a mission-critical issue in another platform 1, while continuous cooperation and corroboration of Platform 3 and corroboration of HITL via security cameras, demonstrated and validated the increased autonomous mission resilience via the symbiotic interactions. The increased symbiosis will allow the development of more complex multi-robot autonomous missions in the offshore and various other industrial sectors, improving the efficiency of the inspection and maintenance.

## VI. CONCLUSION

In this paper we described and evaluated our SSOSA for improving the resilience of MR-fleets. We have seamlessly integrated the telemetry data from various robotic platforms into our digital twin, which has allowed the HITL to corroborate and trigger pre-determined multi-robot autonomous missions. Our symbiotic architecture demonstrates bidirectional capabilities, which advance interaction across a multi-robot fleet. This leverages the advantages of our SSOSA to meet the resilience and safety compliance demands of multi sector operating environments. This represents a departure from current product-centric designs allowing the first steps toward a bio-inspired multi-robot fleet. A limitation of product centric design includes low level interactions across cooperation, collaboration and corroboration elements in single agent autonomous swarms. A potential solution to this limitation includes the application of a digital twin, which can interact with heterogeneous multi-robot fleets via bio-inspired symbiosis.

This paper has demonstrated that bio-inspired symbiosis in a digital environment can enhance resilience in a multi-robot autonomous mission. Three missions with an increasing number of inter-robot interactions, via $C^3$ governance, have validated the improved mission resilience in the multi-robot autonomous missions. Within M3, AI models were trained to identify a robot (platform 1) and representation of a battery box to achieve the mission goal.

Introducing further symbiotic features in the multi-robot missions and collecting the data during each mission will enable an ongoing machine learning strategy for these systems. Thus, enhancing mission resilience to create a self-sustainable multi-robot autonomous system.

## ACKNOWLEDGEMENTS

The research within this paper has been supported by the Offshore Robotics for Certification of Assets (ORCA) Hub [EP/R026173/1].